\documentclass{article}
\usepackage{spconf,amsmath,epsfig}
\usepackage{comment}

\usepackage{amssymb}
\usepackage{booktabs}       
\usepackage{threeparttable}  
\usepackage{paralist}
\usepackage{footnote}
\usepackage{tablefootnote}
\usepackage{verbatim}
\usepackage{comment}
\usepackage{subfig}
\usepackage{caption}
\PassOptionsToPackage{hyphens}{url}
\usepackage{url} 
\usepackage{algorithm}

\title{Road Network and Travel Time Extraction from Multiple Look Angles with SpaceNet Data}
\name{Adam Van Etten, Jacob Shermeyer, Daniel Hogan, Nicholas Weir, Ryan Lewis}
\vspace{-5pt}
\address{In-Q-Tel CosmiQ Works \\
    {\tt\small \{avanetten, jshermeyer, dhogan, nweir, rlewis\}@iqt.org}
}

\begin{document}
%
\maketitle
\begin{abstract}

Identification of road networks and optimal routes directly from remote sensing is of critical importance to a broad array of humanitarian and commercial applications. Yet while identification of road pixels has been attempted before, estimation of route travel times from overhead imagery remains a novel problem, particularly for off-nadir overhead imagery. 
To this end, we extract road networks with travel time estimates from the SpaceNet 
MVOI
dataset. Utilizing the
CRESIv2
framework, we demonstrate the ability to extract road networks in various observation angles
and quantify performance at 27 unique nadir angles 
with the graph-theoretic 
APLS$_{\rm{length}}$ and APLS$_{\rm{time}}$ metrics.  
A minimal gap of 0.03 between APLS$_{\rm{length}}$ and APLS$_{\rm{time}}$ scores indicates that our approach yields speed limits and travel times with very high fidelity.
We also explore the utility of incorporating all available angles during model training, and find a peak score of APLS$_{\rm{time}} = 0.56$.  The combined model exhibits greatly improved robustness over angle-specific models, despite the very different appearance of road networks at extremely oblique off-nadir angles versus images captured from directly overhead.


\end{abstract}

\section{Introduction}
\label{sec:intro}

Using convolutional neural networks to interpret overhead imagery has applications in disaster response \cite{fb, damage_assessment}, agriculture \cite{ag2}, and many other domains \cite{remote_sensing}.  Remote sensing satellites with high spatial resolution often have to point their sensors off-nadir to capture areas of interest if they are not 
directly overhead.
This is particularly common in real-world use cases when timely collection and analysis is required, necessitating collection from an oblique (off-nadir) look angle.  This motivates an analysis of how viewing angle affects deep learning model performance.  
The effect of viewing angle on model performance for finding building footprints has been previously studied \cite{mvoi},
but here an analogous study for roads is undertaken for the first time.  The ability to construct a road network from a satellite image is representative of a broad class of geospatial deep learning problems, while also being intrinsically valuable for routing during rapidly-changing conditions.  Key to this capability is going beyond pixel segmentation to extracting and evaluating a graph-theoretic representation of a road network.

This paper is organized as follows: Section \ref{sec:dataset} describes related work and the dataset used in this study, while Section \ref{sec:algo} details our algorithmic approach.  Section \ref{sec:experiments} discusses our experiments, and Section \ref{sec:results} details the results of these experiments.  Finally, in Section \ref{sec:conclusions} we discuss takeaways from this paper and the relative success of our efforts.

\section{Related Work and Dataset}
\label{sec:dataset}

Extensive computer vision research has focused on natural scenes, and although there exist natural scene datasets with multiple views (e.g., \cite{cityscapes}) or even 3D models (e.g., \cite{scans}), neither category encompasses a wide range of viewing angles in a photorealistic way.  Likewise, a variety of overhead imagery datasets are available, but generally with each area of interest seen from only one or a limited range of views (e.g., \cite{isprs}).
To evaluate algorithmic ability to extract routable road networks from satellite imagery we turned to the SpaceNet Multi-View Overhead Imagery (MVOI) dataset \cite{mvoi}. To our knowledge, SpaceNet MVOI is the only open access high-resolution satellite imagery dataset with a wide range of viewing angles of the same geographic area and time. 

SpaceNet MVOI comprises 27 distinct collects acquired in a 5 minute time span during a single pass of the Maxar WorldView-2 satellite over Atlanta. These looks range from $7^{\circ}$ to $53^{\circ}$ off-nadir, including images both before and after the satellite passed over Atlanta (i.e. target azimuth angle $17.7^{\circ}$ to $182.8^{\circ}$), all covering the same 665 km$^{2}$ of geographic area. 
As look angle increases from $7^{\circ}$ to $53^{\circ}$ and images cover progressively larger ground areas, 
the pixel size (ground sample distance, GSD) also increases from 0.5 m to 1.67 m. 
Alongside the imagery, MVOI includes manually curated labels for machine learning: 126,747 building footprint polygons and $\approx3,000$ km of road network linestrings.  The road network labels contain metadata indicating number of lanes, road type (residential surface road, major highway, etc.) and surface type.  These attributes dictate estimated safe travel speed; for example, a paved one-lane residential road has a speed limit of 25 mph, while a three-lane paved motorway can be traversed at 65 mph, and a one-lane dirt cart track has a traversal speed of 15 mph \cite{cresiv2}. Both building and road network labels were manually annotated using the most nadir ($7^{\circ}$) collect.  The images and labels are tiled into $450 \rm{m} \times 450 \rm{m}$ ($0.2025 \, \rm{km}^{2}$) chips for machine learning. 80\% of the geographic tiles are included in the training set and 20\% are held back for testing (yielding 231 unique testing scenes at each angle), with each set comprising all 27 collects at their respective geographic tiles. For further details regarding the dataset see \cite{mvoi} or \url{www.spacenet.ai}.  

\section{Algorithmic Approach}
\label{sec:algo}

\begin{figure}[]
\vspace{-1pt}
\begin{center}
\centering
\setlength{\tabcolsep}{0.3em}
\begin{tabular}{cc}
\vspace{-1pt}
\subfloat [\textbf{Training Image}] {\includegraphics[width=0.43\linewidth]{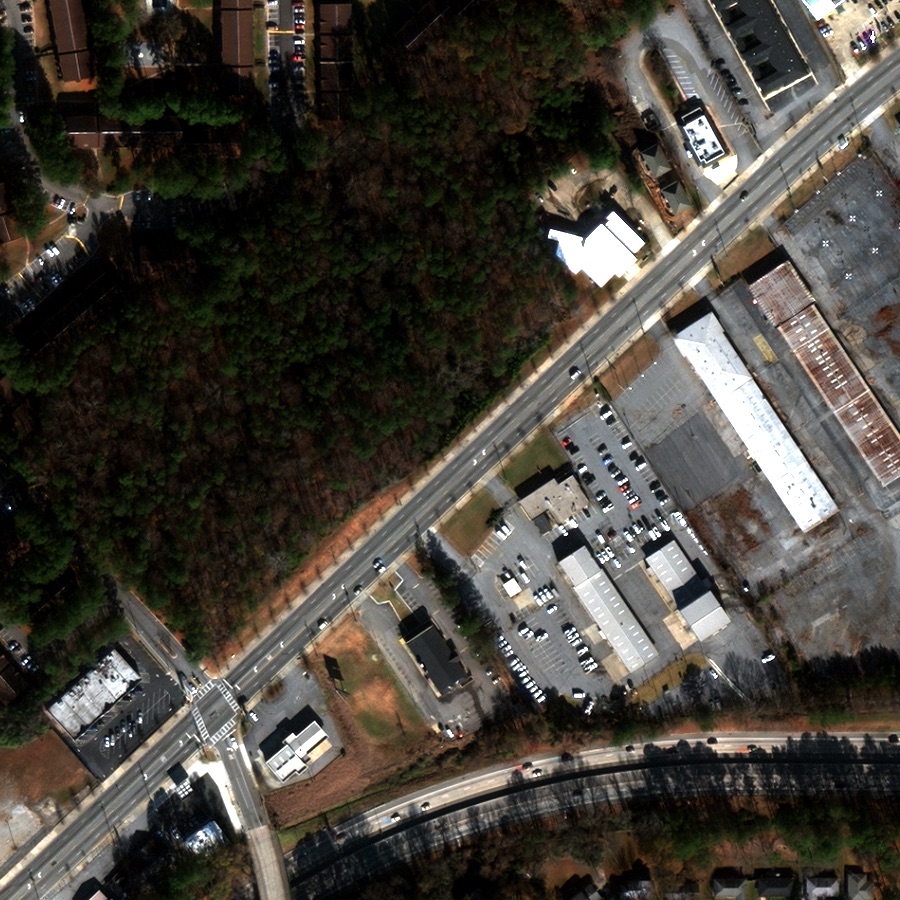}} &
\subfloat [\textbf{Multi-channel Mask}] {\includegraphics[width=0.43\linewidth]{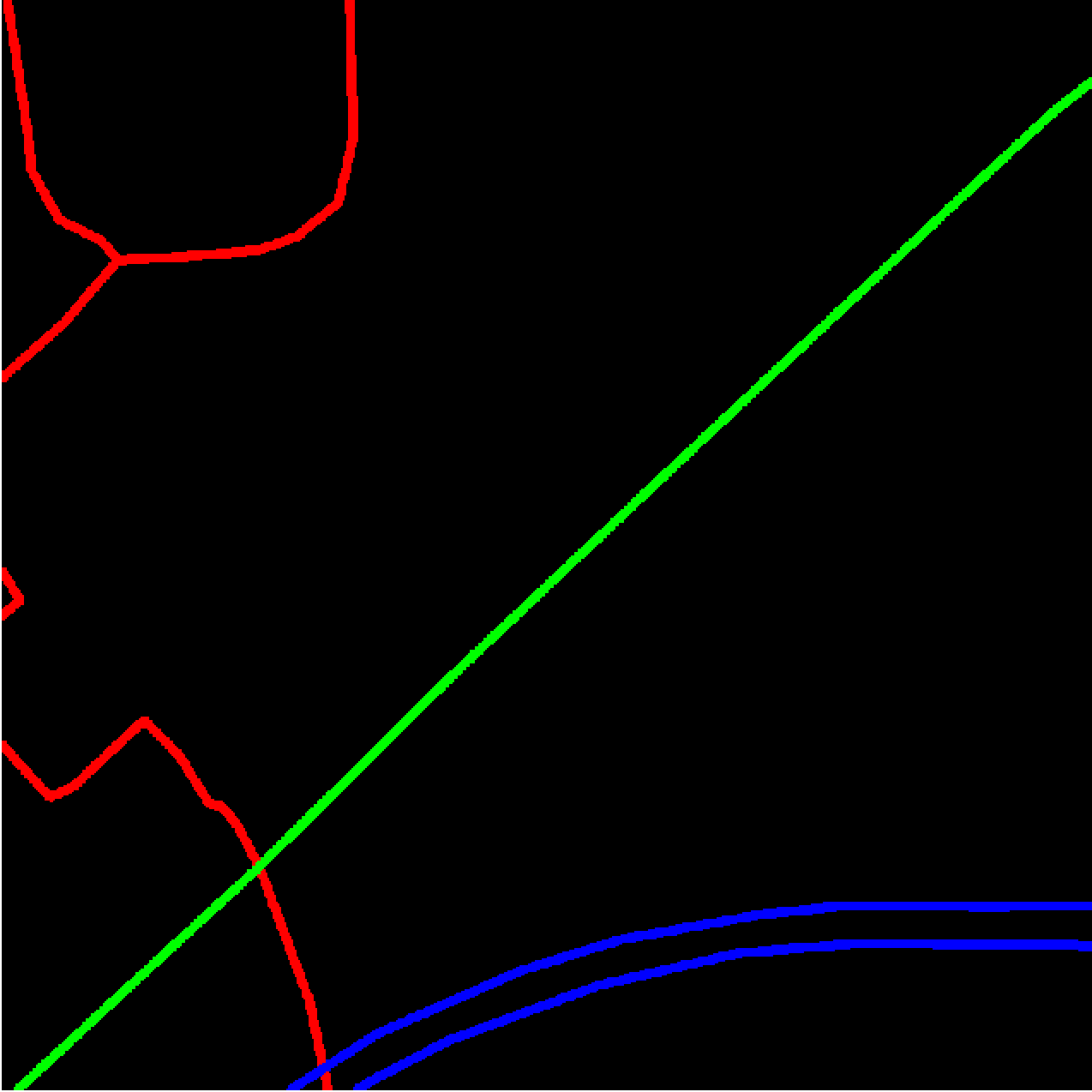}} 
\end{tabular}
 \caption{\textbf{Sample training data.} (a) Example Atlanta training chip. (b) Multi-channel training mask with roads colored by speed (red = 25 mph, green = 45 mph, blue = 55 mph). 
  }
 \label{fig:drp}
\end{center}
\vspace{-12pt}
\end{figure}

\begin{figure}[]
\vspace{-1pt}
  \centering
     \includegraphics[width=0.9999\linewidth]{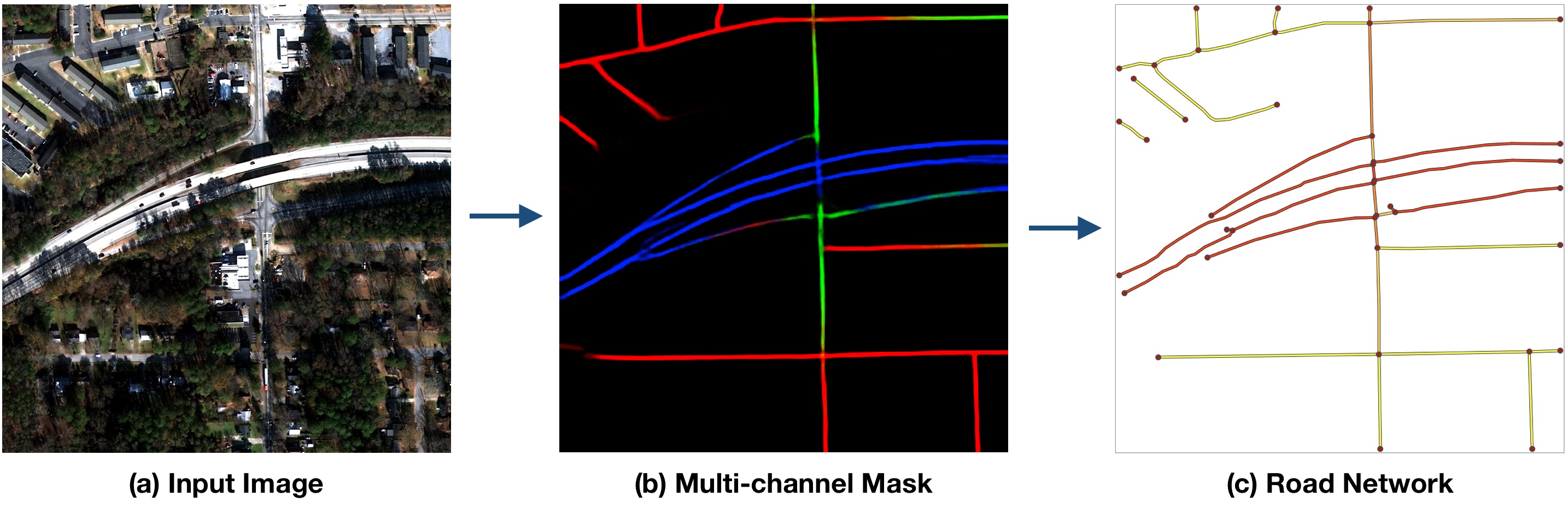}
  \caption{\textbf{CRESIv2 road extraction algorithm.} This schematic collapses the multiple phases of CRESIv2 into two bins: image segmentation and refinement (a$\xrightarrow{}$b), and graph extraction with speed inference (b$\xrightarrow{}$c). The output network (c) is colored by speed from yellow (25 mph) to red (65 mph).
  }
   \label{fig:cresi_algo}
 \vspace{-1pt}
\end{figure}

We utilize the open source City-scale Road Extraction from Satellite Imagery v2 (CRESIv2) algorithm \cite{cresiv2} that served as the baseline for the recent SpaceNet 5 Challenge focused on road networks and optimized routing from satellite imagery \cite{sn5}.  We render road labels into a multi-channel training mask (Figure \ref{fig:drp}), and train a deep learning segmentation model (ResNet34 \cite{resnet} encoder and a U-Net \cite{unet} inspired decoder)
with these  masks.  At inference time the predicted masks are refined, smoothed, and transformed into a skeleton.  A graph structure is built from this skeleton, with nodes representing intersections and weighted edges representing roads with safe travel speed estimates.  
The channels of the prediction mask provide speed estimates (e.g. channel 2 corresponds to 25 mph), which in turn yield travel times for each roadway. Figure \ref{fig:cresi_algo} summarizes the algorithmic approach.


Inference runs at 280 km$^2$ / hour using a single Titan X GPU. Though CRESIv2-generated labels do not achieve the same fidelity as human labelers, this rate nonetheless represents a significant acceleration over existing manual techniques.  For example, at this speed a 4-GPU cluster could map the entire 9100 km$^2$ area of Puerto Rico in $\approx8$ hours, far faster than the $>2$ months required to manually re-map Puerto Rico after Hurricane Maria \cite{osm_maria}.

\section{Experiments}
\label{sec:experiments}

We begin by training a model solely on the most nadir ($7^{\circ}$) imagery used for labeling. 
We then evaluate this model over all 27 angles in the test dataset (see Section \ref{sec:results} for results).

We also train models by combining all data within a certain angle range.  We use four nadir bins: 
NADIR ($\leq25^{\circ}$, OFF ($26-39^{\circ}$), VOFF ($\geq40^{\circ}$), and ALL.
Each model is trained for 3 days on a single NVIDIA Titan X GPU, which equates to $15 - 65$ epochs, depending on bin size. 

Scoring is accomplished via the graph-theoretic Average Path Length Similarity (APLS) metric \cite{sn3}. This metric sums the differences in optimal path lengths between nodes in the ground truth graph G and the proposal graph G'. The definition of shortest path can be user defined; we focus on APLS$_{\rm{time}}$ metric \cite{cresiv2} to measure differences in travel times between ground truth and proposal graphs, but also consider geographic distance as the measure of path length (APLS$_{\rm{length}}$).

\section{Results}
\label{sec:results}

For the $7^{\circ}$ model, performance at close to nadir angles peaks at APLS$_{\rm{time}} = 0.57$.  For comparison, the highest score posted on a single city by the winner of the SpaceNet 5 challenge was 0.51 \cite{sn5winners}. 
We observe near equal performance when weighting edges with length or travel time, with an APLS$_{\rm{length}}$ score only 0.03 
higher than APLS$_{\rm{time}}$.  These results indicate that road speeds and travel times are extracted quite precisely, as any error in travel time would compound existing errors in the road network topology.

\begin{figure*}[h]
\vspace{-1pt}
\begin{center}
\centering
\setlength{\tabcolsep}{0.3em}
\vspace{-1pt}
\subfloat {\includegraphics[width=0.99\linewidth]{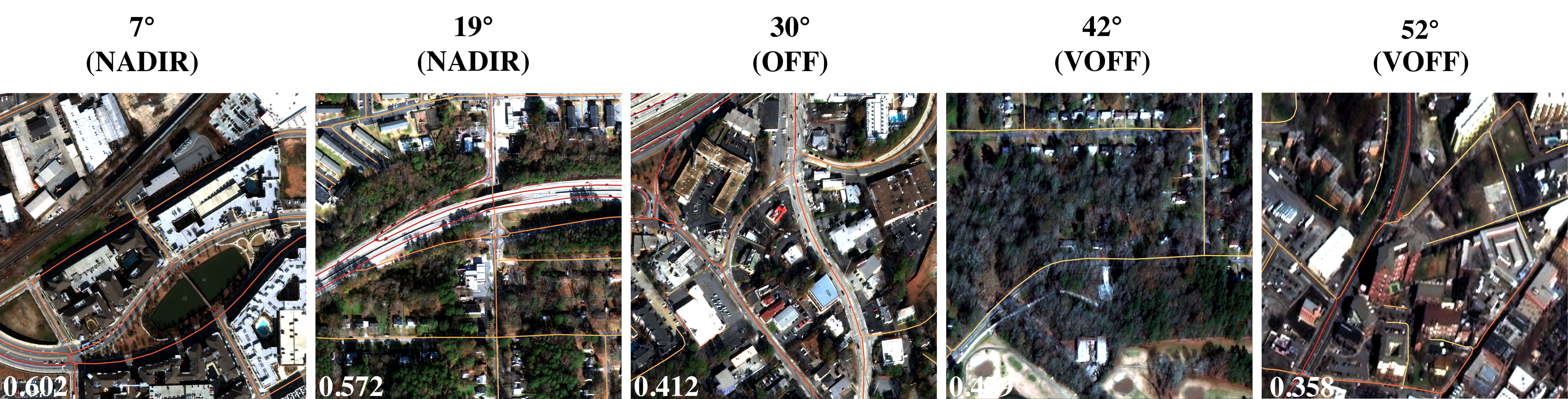}}\\
\caption{\textbf{Speed inference results, `ALL' model.}  Roads are colored by speed, from yellow (25 mph) to red (65 mph).  APLS$_{\rm{time}}$ scores are displayed in the bottom left corner of each chip.
}

\label{fig:grid_speed}
\end{center}
\vspace{-10pt}
\end{figure*}

\begin{figure}[h!]
\vspace{-1pt}
\begin{center}
\centering
\setlength{\tabcolsep}{0.3em}
\vspace{-1pt}
\subfloat {\includegraphics[width=0.99\linewidth]{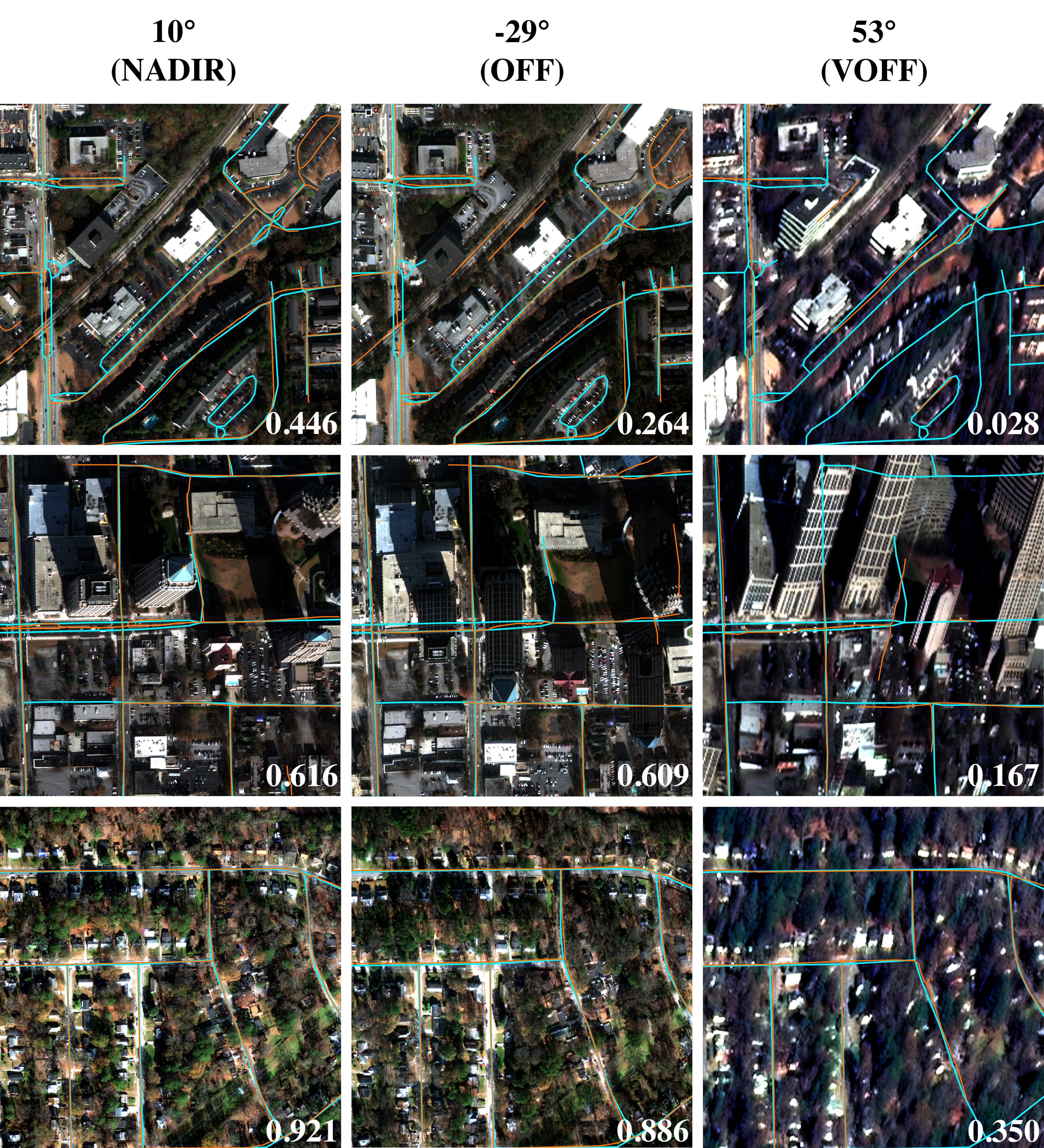}}\\
\caption{\textbf{Results at different look angles, `ALL' model.}  Each row is the same geographic chip, with each column a unique observation angle. Ground truth labels are colored in cyan, with model predictions in orange. APLS$_{\rm{time}}$ scores are displayed in the bottom right of each chip.
}

\label{fig:grid}
\end{center}
\vspace{-12pt}
\end{figure}

Figure \ref{fig:grid_speed} displays the inferred speed for a selection of test chips, with successful differentiation of speed for different road types. 
In Figure \ref{fig:grid} we show predictions and ground truth for multiple nadir angles and test chips. Note that the algorithm frequently successfully connects roads even when overhanging trees obscure the road.  The model also has some limited success in connecting occluded roads behind buildings. 

\begin{figure}[h!]
\vspace{-1pt}
\begin{center}
\centering
\setlength{\tabcolsep}{0.3em}
\begin{tabular}{cc}
\vspace{-1pt}
\subfloat {\includegraphics[width=0.99\linewidth]{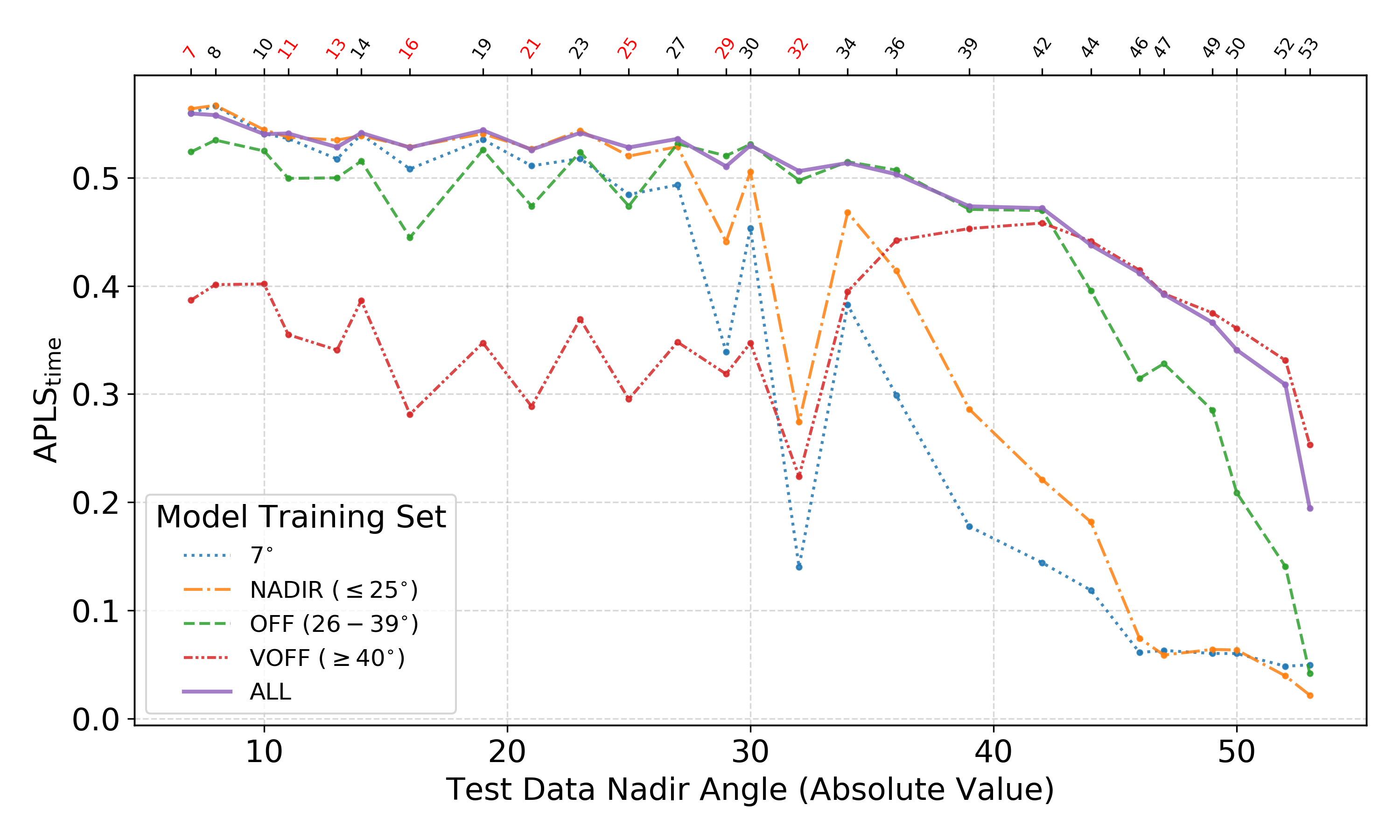}}\\
\vspace{-15pt}\\
\subfloat {\includegraphics[width=0.99\linewidth]{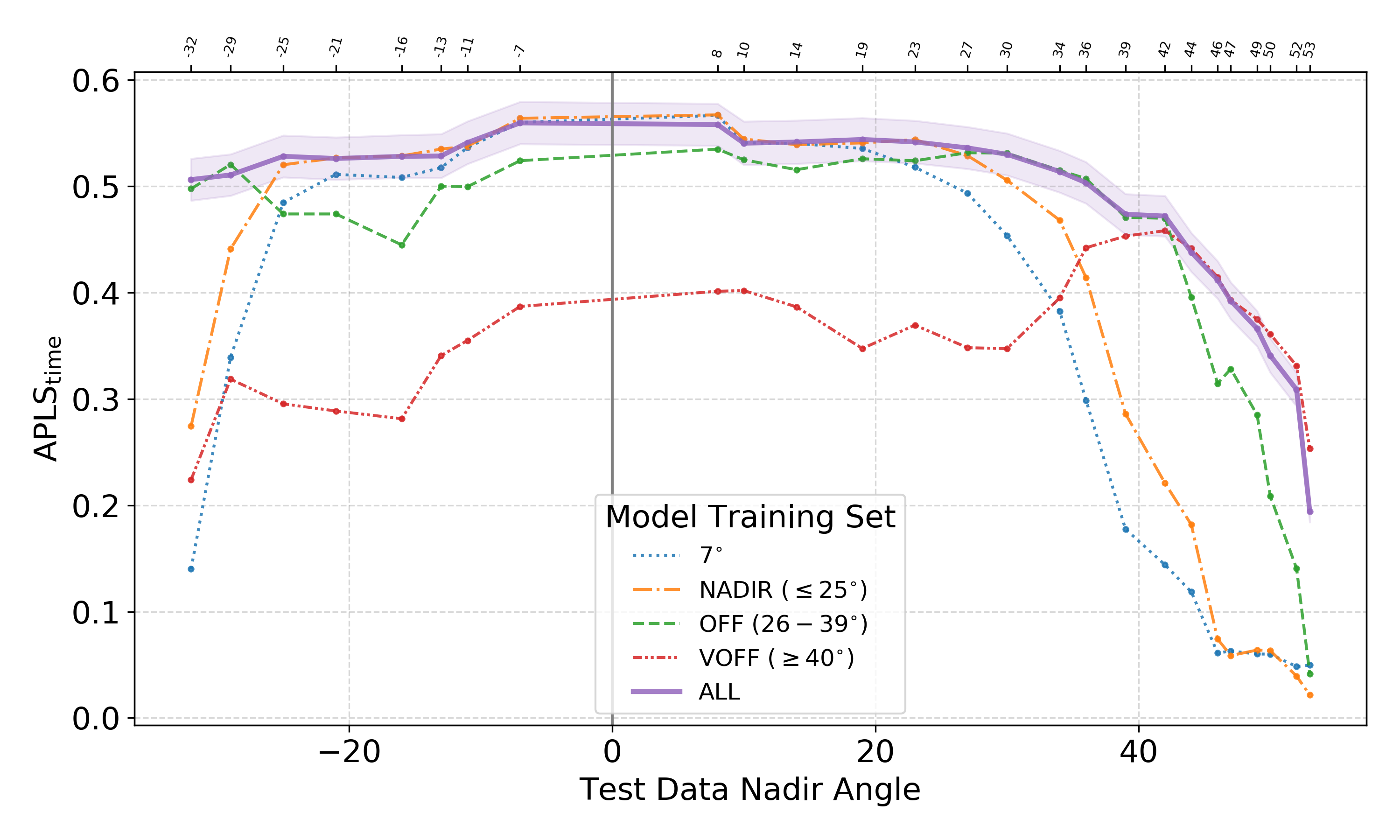}}\\
\end{tabular}
\caption{\textbf{Results for each model.}  APLS$_{\rm{time}}$ scores for at each nadir angle for each of the five trained models.
{Top:} The x-axis is the absolute value of the nadir angle (south facing angles in red). Note the sawtooth pattern caused by these south facing angles.
{Bottom:} The same information as above, but with south facing angles plotted as negative; we also plot the standard error of the mean for the combined model. 
}
\label{fig:glob_res}
\end{center}
\vspace{-10pt}
\end{figure}

We plot the results for each model in Figure \ref{fig:glob_res},  (e.g. the `NADIR' model is trained on data with angles from $7-25^{\circ}$).  The 0.03 difference between APLS$_{\rm{time}}$ and APLS$_{\rm{length}}$ that we noted 
for the 7$^{\circ}$ model
is mirrored for all other models, so for clarity we only plot APLS$_{\rm{time}}$.
In Figure \ref{fig:glob_res} (Top) the x-axis is the absolute value of the nadir angle; of particular note is the pronounced sawtooth pattern for 
all models,
which was also observed for building footprint extraction in \cite{mvoi}.  The dips correspond to south-facing shots looking into the sun where we hypothesize that shadows complicate road  extraction.

The bottom panel of Figure \ref{fig:glob_res} renders south-facing nadir angles as negative, and illustrates that each model performs well in the angle bin it was trained in. Of particular import is that the model trained on all bins (i.e. `ALL') equals performance (within errors) of the bin-specific (e.g. `VOFF') models.  Evidently, the `ALL' model incorporating all training angles is far more robust than bin-specific models, with (APLS$_{\rm{time}} > 0.5$) for nadir angles between -32 and +36 degrees.  In the very off-nadir bin of 40 degrees or greater we observe a marked drop in performance,  with APLS$_{\rm{time}} \approx 0.2$ at the highest nadir angle of $53^{\circ}$. 

\section{Conclusions}
\label{sec:conclusions}

We utilize the heretofore unexplored SpaceNet MVOI road labels to train models for road network and travel time extraction at both on- and off-nadir imagery.
For a model trained solely on a single nadir collect (taken at a mere $7^{\circ}$ off-nadir) we achieve reasonable APLS scores out to $\approx25^{\circ}$ off-nadir, though at higher nadir angles performance with this model drops precipitously.   We find that incorporating all available data regardless of inclination angle into one model is far more robust than bin-specific models trained on a subset of look angles.  This global model achieves scores of APLS$_{\rm{time}} > 0.5$) for nadir angles between -32 and +36 degrees, though road network inference at very high off-nadir angles of $\geq45^{\circ}$ is extremely challenging.

Road network extraction performance at off-nadir angles has a somewhat different functional form than building extraction at off-nadir.  Comparing Figure \ref{fig:glob_res} to \cite{mvoi}, we note that for buildings the bin-specific model outperforms the global model in the very off-nadir regime; yet for roads we find that a global model performs well at all angles.  For roads, we observe a $65\%$ drop in performance between nadir and $53^{\circ}$ off-nadir; contrast this to buildings, where published results indicate a $91\%$ drop in score between nadir and $53^{\circ}$ off-nadir.  It appears that inferring occluded roads in high off-nadir shots is easier than inferring building footprints.  This may be due in part to the greater utility that context plays for roads; since roads are usually connected, surrounding roadways are able to inform occluded roads, while surrounding buildings yield less information about occluded buildings. 

Surprisingly, the APLS$_{\rm{time}}$ and APLS$_{\rm{length}}$ scores are nearly identical across all look angles ($\sim6\%$ average difference), despite the additional requirement to extract safe travel speed for estimating APLS$_{\rm{time}}$. This suggests that CRESIv2 can estimate safe travel speed nearly perfectly, as any other result would have further reduced the APLS$_{\rm{time}}$ score by compounding existing errors associated with route length. As our estimate for safe travel speed is defined by road size, surface type, and context (e.g. residential road vs. major highway), this implies that CRESIv2 learns road attributes as well as their layout. 

Automated extraction of road speeds and travel times from off-nadir satellite imagery applies to a great many problems in the humanitarian and disaster response domains; this paper has demonstrated that such a task is not only possible, but available in the open source and far faster than manual annotation.





{\small
\bibliographystyle{IEEEbib}
\bibliography{strings,refs}
}
\end{document}